\documentclass[runningheads]{llncs}
\usepackage[T1]{fontenc}
\usepackage{graphicx}
\usepackage{booktabs}
\usepackage[misc]{ifsym}
\usepackage{algorithm}
\usepackage{algorithmic}
\usepackage{subfigure}
\usepackage{multicol}
\usepackage{multirow}
\usepackage[switch]{lineno}
\usepackage{float}
\usepackage{xcolor}
\usepackage{amssymb}
\newcommand{\corr}{(\Letter)}
\usepackage{mwe}

\begin{document}

\title{GraphRPM: Risk Pattern Mining on Industrial Large Attributed Graphs}



\author{Sheng Tian\thanks{Co-first authors with equal contributions}\inst{1} \and Xintan Zeng$^\star$\inst{1} \and Yifei Hu$^\star$\inst{1} \and Baokun Wang\inst{1}\corr \and Yongchao Liu\inst{1}\corr \and Yue Jin\inst{1} \and Changhua Meng\inst{1} \and Chuntao Hong\inst{1} \and Tianyi Zhang\inst{1} \and Weiqiang Wang\inst{1}}

\authorrunning{Tian et al.}

\institute{Ant Group, China
\email{\{tiansheng.ts,xintan.zxt,liuxu.hyf,yike.wbk,yongchao.ly,jinyue.jy,\\changhua.mch,chuntao.hct,zty113091,weiqiang.wwq\}@antgroup.com}
}

\toctitle{GraphRPM: Risk Pattern Mining on Industrial Large Attributed Graphs}
\tocauthor{Sheng~Tian,Xintan~Zeng,Yifei~Hu,Baokun~Wang,Yongchao~Liu,Yue~Jin,Changhua~Meng,Chuntao~Hong,Tianyi~Zhang,Weiqiang~Wang}

\maketitle              

\begin{abstract}
Graph-based patterns are extensively employed and favored by practitioners within industrial companies due to their capacity to represent the behavioral attributes and topological relationships among users, thereby offering enhanced interpretability in comparison to black-box models commonly utilized for classification and recognition tasks. For instance, within the scenario of transaction risk management, a graph pattern that is characteristic of a particular risk category can be readily employed to discern transactions fraught with risk, delineate networks of criminal activity, or investigate the methodologies employed by fraudsters. Nonetheless, graph data in industrial settings is often characterized by its massive scale, encompassing data sets with millions or even billions of nodes, making the manual extraction of graph patterns not only labor-intensive but also necessitating specialized knowledge in particular domains of risk. Moreover, existing methodologies for mining graph patterns encounter significant obstacles when tasked with analyzing large-scale attributed graphs. In this work, we introduce GraphRPM, an industry-purpose parallel and distributed risk pattern mining framework on large attributed graphs. The framework incorporates a novel edge-involved graph isomorphism network (EGIN) alongside optimized operations for parallel graph computation, which collectively contribute to a considerable reduction in computational complexity and resource expenditure. Moreover, the intelligent filtration of efficacious risky graph patterns is facilitated by the proposed evaluation metrics. Comprehensive experimental evaluations conducted on real-world datasets of varying sizes substantiate the capability of GraphRPM to adeptly address the challenges inherent in mining patterns from large-scale industrial attributed graphs, thereby underscoring its substantial value for industrial deployment.

\end{abstract}

\section{Introduction}

\begin{figure}[t!]
  \centering
  \includegraphics[width=0.8\linewidth]{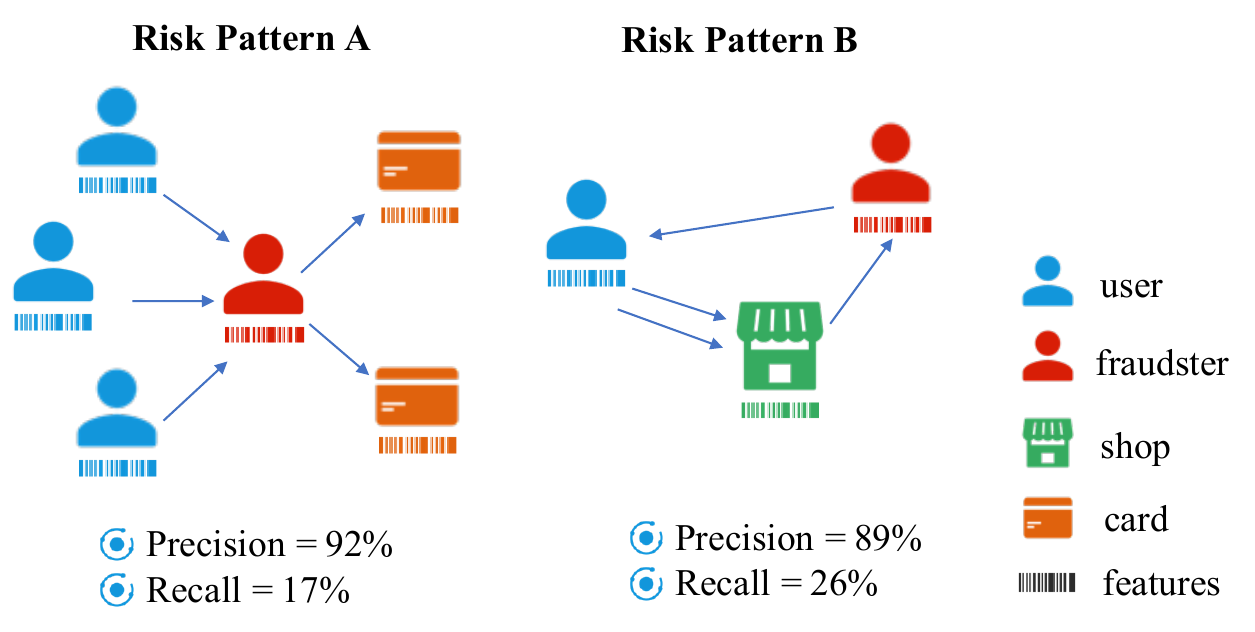}
  \caption{Example risk patterns. Risk pattern A describes the behavior of fraudsters who defraud funds from multiple victim users and quickly transfer them to different downstream bank cards. Risk pattern B describes that the fraudster collects the victim's funds multiple times in the name of investment through a shop, giving rewards in the early stage but no longer paying in the later stage. Precision and recall metrics are the evaluation criteria for measuring risk patterns in the industry.}
  \label{fig.risk_pattern}
\end{figure}

Graph pattern mining constitutes a pivotal task within the ambit of mining and machine learning, with profound applications extending to various industrial and business domains such as social network analysis~\cite{Community2017Seyed}, financial fraud detection~\cite{Migration2009xiaoxi,liu2022eriskcom}, and computational bioinformatics~\cite{Efficiently2005Mohammed}. Taking the financial transaction scenario as an example, fraudsters would try to cheat normal users and make illegal money transfers. The distinctive behavioral patterns of these fraudsters, termed 'risk patterns', are critical for the detection of fraudulent activity and the prevention of financial fraud, as exemplified in Fig.~\ref{fig.risk_pattern}. Compared to black-box neural network models used for identifying fraudsters~\cite{luo2023faf}, industry experts express a preference for summarizing these risk patterns, as they provide more granular insight into the conduct of fraudulent entities, thereby facilitating a more explainable approach to fraud detection. Nonetheless, the manual delineation or construction of these patterns by experts is a labor-intensive process that demands considerable domain-specific knowledge. Consequently, the automation of risk graph pattern mining is an avenue warranting exploration. GRAMI~\cite{GRAMI2014Mohammed} presents a method for frequent subgraph mining by leveraging a novel canonical labeling technique to efficiently discover patterns within a single large graph. Bliss~\cite{bliss2007} introduces an optimized tool for canonical labeling, specifically designed to handle the challenges posed by large and sparse graph structures, enhancing the performance of graph mining tasks. T-FSM~\cite{tfsm2023yuan} outlines a task-based framework that enables massively parallel processing for frequent subgraph pattern mining, addressing the scalability issues associated with big graph data. Despite this interest, extant automated graph pattern mining algorithms~\cite{GRAMI2014Mohammed,bliss2007,Subgraph2022Nguyen,tfsm2023yuan} are impeded by two principal limitations:
\begin{enumerate}
    \item \textbf{Challenges in processing attributed graphs}. In numerous real-world applications, simplistic representations of graph topology fall short of accurately depicting risk scenarios. There is a necessity to leverage high-dimensional attributes associated with nodes or edges for a nuanced characterization of entities, which is beyond the capabilities of methods that are restricted to or can only process one-dimensional attributes.
    
    \item \textbf{Deficiencies in scalability}. Graph data within industrial environments is characteristically voluminous, spanning millions or even billions of nodes. Existing methodologies lack the integration of computational optimization strategies that are critical for the effective and efficient management of data at such an industrial scale. This shortfall in capability significantly undermines the suitability of these methods for application in industrial tasks, which necessitate robust data manipulation and analytical capacity to handle the sheer volume and complexity of the data involved.
\end{enumerate}

In this paper, we address the problem of\textit{ \underline{R}isk \underline{P}atterns \underline{M}ining on large transaction attributed graphs} (GraphRPM). Although our research is primarily focused on financial fraud detection, the versatility of the proposed framework allows for its extension to a multitude of industrial applications, including but not limited to analysis within social network contexts. The challenge of managing and processing large-scale attributed graphs in industrial settings is a nontrivial hurdle, particularly in the realm of data mining. The primary objective of this study is to establish a robust and efficacious methodological framework capable of discerning distinct graph patterns as discriminative entities, enabling the differentiation of various graphical structures and the identification of fraud risk patterns.

GraphRPM introduces a pioneering Edge-Involved Graph Isomorphism Network (EGIN) that addresses the challenge of fuzzy matching in attributed graph patterns, striking a balance between computational complexity and accuracy. Furthermore, this study implements a two-stage mining strategy coupled with a parallel distributed processing framework to diminish computational redundancy and enhance efficiency. Additionally, we present a \textit{Pattern Risk Score} as an evaluative measure for identifying salient risk patterns. Comprehensive evaluations across diverse real-world datasets, varying in size, corroborate GraphRPM's proficiency in resolving pattern mining issues within expansive industrial attributed graphs. Our research represents a significant advancement in the application of data mining and machine learning to industrial and business analytics. We contribute to the field in two pivotal ways.
\begin{enumerate}
    \item We meticulously conceptualize and address the hitherto underexplored issue of discerning risk patterns on large-scale attributed graphs.
    \item We introduce an all-encompassing analytical framework that not only incorporates the cutting-edge EGIN algorithm but also integrates a scalable distributed computation system, thereby enhancing computational efficacy. To our knowledge, this is the first proposition of an approximation algorithm based on graph neural networks for risk pattern mining on large transaction-attributed graphs.
\end{enumerate}
\section{Problem Formulation}
The large transaction-attributed graph can be represented as $\mathcal{G}=(\mathcal{V}, \mathcal{E})$, where $\mathcal{V} = \{v_1, v_2,...v_N\}$ is the set of nodes indicating the user, and $\mathcal{E}$ denotes the set of transaction events. Typically, let $e_{ij} \in \mathcal{E}$ be an interaction or a link that happens from source node $v_i$ to target node $v_j$, associated with edge feature $f_e$ and node feature $f_v$, respectively. $f_v$ (and $f_e$) is defined as a feature vector $\mathbf{x}_v \in \mathbb{R}^{d_v}$ (and $\mathbf{x}_e \in \mathbb{R}^{d_e}$), where $d_v$ (and $d_e$) is the dimension size. Note that $f_v$ and $f_e$ often have different dimension sizes in practice.

The risk graph pattern is defined as $P = (g, u)$, consisting of a subgraph $g=(\mathcal{V}_g, \mathcal{E}_g)$ and the starting node $u$ of the pattern. For any node $v_i \in \mathcal{V}$, the node $v_i$ hits the pattern $P$ if $f_{v_i} = f_u$  and $g \in \mathcal{G}_{v_i}$, where $\mathcal{G}_{v_i}$ is the $k$-hop subgraphs of the node $v_i$. For a given node $v_i$, its $k$-hop subgraph compirses all of the nodes that can be reached from node $v_i$ by traversing at most $k$ edges, including the starting node $v_i$ by default unless otherwise specified, and all of the edges that connect the nodes within this defined $k$-hop neighborhood.

We employ the support metric, which is traditionally used in frequent pattern mining~\cite{yan2002gspan,Output2009Mohammad,teixeira2015arabesque,talukder2016a} to denote the prevalence of particular patterns within a dataset. Different from the objectives of frequent pattern mining, our research is focused on the extraction of risk patterns characterized by high support in anomalous instances juxtaposed with low support in normal instances. The goal of our research is to leverage these discriminative patterns to differentiate between normal and abnormal instances within extensive attributed graphs.

\begin{figure}[t!]
  \centering
  \includegraphics[width=\linewidth]{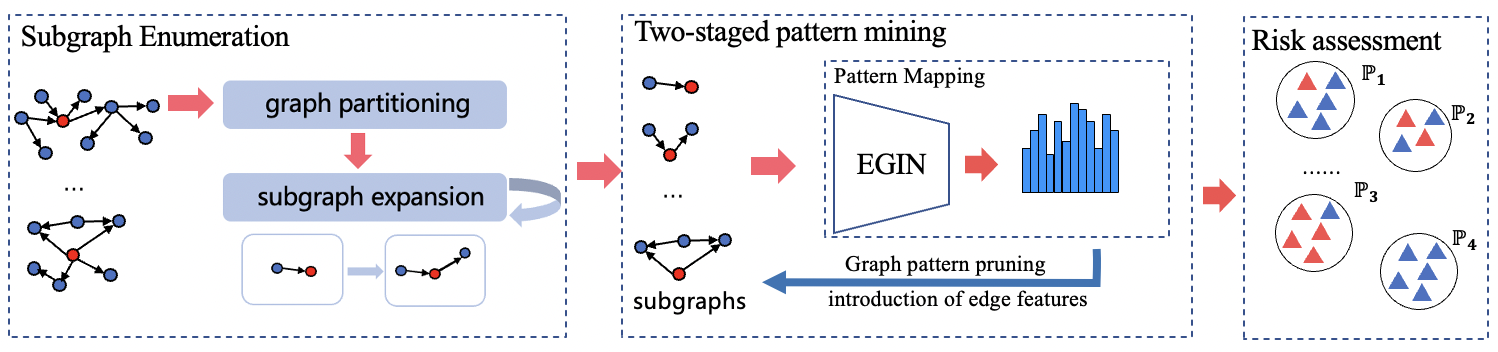}
  \caption{An overview of the GraphRPM framework, which consists of subgraph enumeration, two-stage mining, and risk pattern assessment. The Edge-Involved Graph Isomorphism Network (EGIN) based pattern representation mapping is used to identify graph patterns with attributes.}
  \label{fig.system-architecture-overview}
  \vspace{-0.5cm}
\end{figure}

\section{METHODS}
Fig.~\ref{fig.system-architecture-overview} illustrates the overall workflow of GraphRPM, including potential subgraph enumeration, two-stage pattern mining, and pattern risk assessment, where the EGIN based pattern representation mapping technique is used to identify graph patterns. Owing to the extensive scale of industrial attributed graphs, it necessitates decomposition into smaller ego-graphs to enable the enumeration of potential sub-graph patterns. This is followed by the extraction of graph pattern results through a two-staged pattern mining approach that leverages node and edge attributes within the EGIN network. Ultimately, risk patterns that demonstrate significant distinctiveness are identified and selected via a thorough risk assessment process.

\subsection{Potential Subgraph Enumeration}
To obtain the risk graph pattern, we first need to enumerate the potential subgraph of patterns around each starting node. However, industrial attributed graphs tend to be too large to be processed directly in memory, so we pre-extract the $k$-hop ego-graph of each node and then enumerate the potential subgraphs within the ego-graph. While the node magnitude is still large, to further improve efficiency, GraphRPM performs enumeration utilizing our distributed in-memory graph intelligent computing system~\cite{liu2022graphtheta}, which can handle both graph computing and graph learning tasks and accommodate multiple programming paradigms including the widely recognized vertex-centric programming model~\cite{gonzalez2012powergraph,lin2018shentu,zhu2016gemini}.

\begin{figure}[t!]
  \centering
  \includegraphics[width=0.7\linewidth]{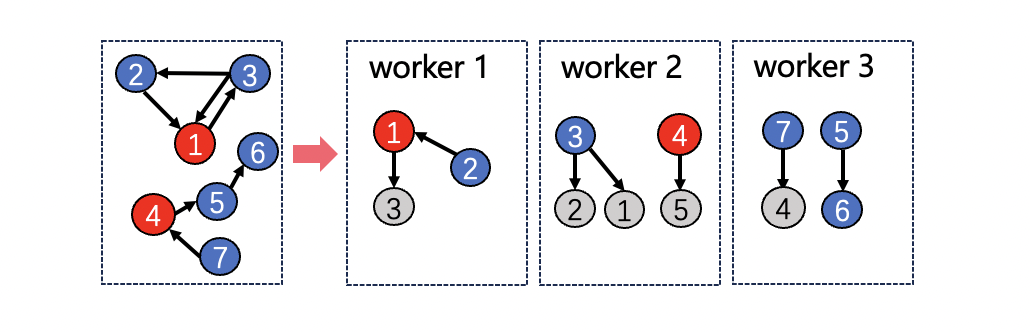}
  \vspace{-0.3cm}
  \caption{An example of graph partitioning, where the nodes are color-coded to indicate their roles. The nodes colored in red serve as the starting nodes, while the nodes colored in blue represent the neighboring nodes of the starting nodes within their respective ego-graphs. To elaborate, $v_1$ and $v_4$ act as the starting nodes. $v_1$'s ego-graph encompasses its neighbor nodes $v_2$ and $v_3$, while $v_4$'s ego-graph includes nodes $v_5$, $v_6$, and $v_7$ as its neighbors. All the aforementioned nodes are designated as master nodes. In addition, nodes colored in gray signify the mirror nodes, which are replicas of the master nodes and carry the same IDs. For instance, the gray node marked $v_3$ within worker $1$ functions as a mirror for the master node $v_3$ located in worker $2$.}
  \label{fig.partitioning}
  \vspace{-0.3cm}
\end{figure}

Each worker holds a partition of the input data and runs on multiple threads. As shown in Fig.~\ref{fig.partitioning}, we illustrate the process of distributing data using a simple graph partitioning algorithm. Initially, all nodes are evenly distributed across the workers, referred to as master nodes. For each edge, our method assigns it to the partition where its source node is a master node. If the target node of these edges does not reside within the same partition, they are created as mirror nodes (represented by grey dots in the figure) within that partition, functioning as replicas of their corresponding master nodes.

\begin{figure}[t!]
  \centering
  \includegraphics[width=0.7\linewidth]{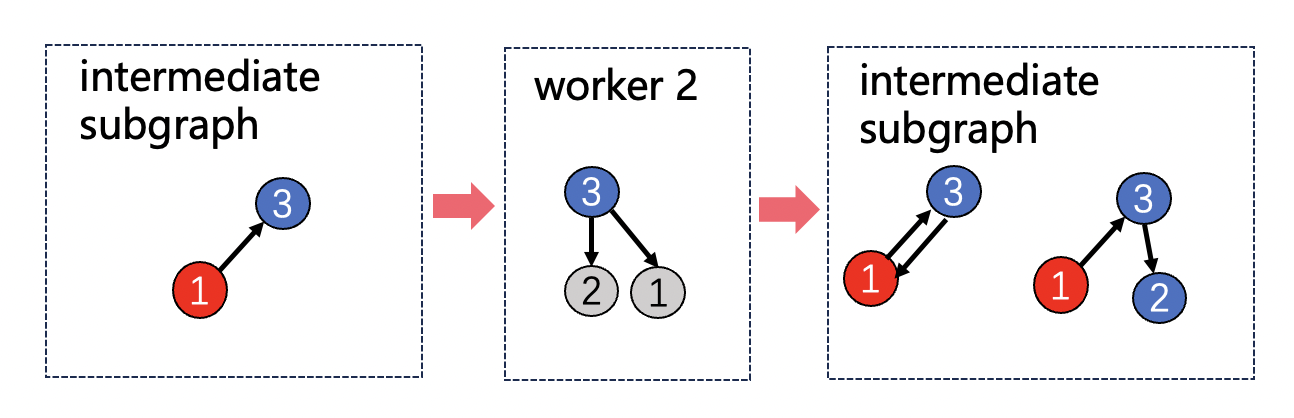}
  \vspace{-0.3cm}
  \caption{An example of expansion. (1) Left: an intermediate subgraph of edge set $\{e_{1,3}\}$ generated during the first iteration of expansion from the node $v_1$ is transmitted to $v_3$. (2) Middle: in the second iteration of expansion, $v_3$ will be activated and adds its edges $e_{3,1}$ and $e_{3,2}$ to this subgraph, respectively, thus producing two new intermediate subgraphs of edge sets $\{e_{1,3}, e_{3,1}\}$ and $\{e_{1, 3}, e_{3,2}\}$. (3) Right: in this case, the two subgraphs will be further transmitted to $v_3$'s neighbor nodes $v_1$ and $v_2$.}
  \label{fig.expansion}
  \vspace{-0.3cm}
\end{figure}

Upon completion of the graph partitioning, the system proceeds to execute the subgraph enumeration algorithm. We use a breadth-first-search (BFS) strategy to fully utilize the parallel computation. The enumeration begins from around each starting node, generating a series of subgraphs containing only a single edge, which are then transmitted to their respective neighboring nodes. As shown in Fig.~\ref{fig.expansion}, an intermediate subgraph of edge set $\{e_{1,3}\}$ generated during the first iteration of expansion from starting node $v_1$ is transmitted to its neighbor node $v_3$. In the subsequent iteration of expansion, node 3 will be activated and adds its edges $e_{3,1}$ and $e_{3,2}$ to this subgraph, respectively, thus producing two new intermediate subgraphs of edge sets $\{e_{1,3}, e_{3,1}\}$ and $\{e_{1, 3}, e_{3,2}\}$. In this case, the two subgraphs will be further transmitted to the two neighbor nodes $v_1$ and $v_2$ of $v_3$.
After each iteration, neighboring nodes will be activated and traverse their edges, attempting to add them into the subgraph. Enumeration stops when either the size of the subgraph reaches a threshold or there are no more active nodes. Obviously, BFS suffers from memory issues due to the maintenance of enormous intermediate data, we introduce several optimization methods as follows.

\subsubsection{Coordination-free Redundant Subgraph Removal}
Two different workers may reach the same subgraph due to different edge-induced orders, causing the subgraph to be enumerated and represented twice. To solve this problem, we develop a coordination-free technique to avoid redundant computation and minimize communication costs. This method enforces restrictions on the order of edge IDs by sorting to form the representation of each subgraph and then applies a hashing technique to the representation by each worker independently to decide which worker should compute the subgraph.

\begin{figure}[t!]
  \centering
  \includegraphics[width=0.6\linewidth]{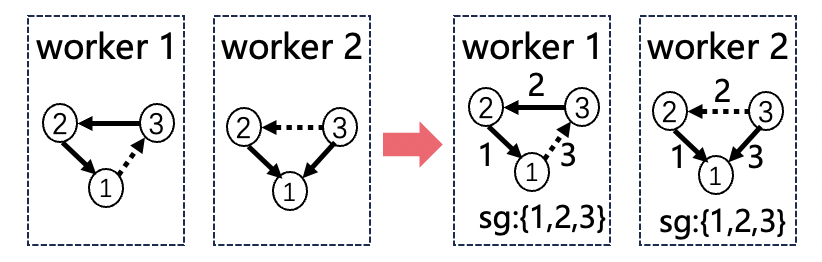}
  \caption{An example of coordination-free technique. (1) Left: two different workers reach the same subgraph. One starts from edges $\{e_{2,1}, e_{3,2}\}$ by adding $e_{1,3}$, the other starts from edge $\{e_{2,1}, e_{3,1}\}$ by adding $e_{3,2}$. (2) Right: we assign an ID attribute to each edge in advance, allowing each subgraph to obtain a simplistic representation based on the ascending order of edge IDs. Identical subgraphs will have the same representation. Each worker calculates a hash value and performs modulus operations on the representation of the subgraph, using the total number of edges within the subgraph as the divisor, in order to identify a specific edge within the subgraph. This particular edge is used to decide which worker will process the subgraph with all other works giving up the task, leading to coordination-free distributed redundant subgraph removal.}
  \label{fig.coordinate_free}
  \vspace{-0.3cm}
\end{figure}

Leveraging the example provided in Fig.~\ref{fig.coordinate_free}, we shall elaborate the workings of our coordination-free approach. As displayed in the left panel of the figure, two distinct workers can independently arrive at the same subgraph: worker 1 begins its traversal from the intermediate subgraph of edge set $\{e_{2,1}, e_{3,2}\}$ and incorporates edge $e_{1,3}$, while worker 2 starts from another intermediate subgraph of edge set $\{e_{2,1}, e_{3,1}\}$ and adds edge $e_{3,2}$. To reconcile these overlapping efforts, we preemptively assign a unique ID to each edge. This labeling method enables every worker to derive a streamlined representation of their respective subgraph by arranging the edge IDs in ascending order. Consequently, any identical subgraphs encountered will share the same representative string. Utilizing this uniform representation, workers can then compute a hash value and performs modulus operations on the representation of the subgraph, using the total number of edges within the subgraph as the divisor. This process is employed to pinpoint a particular edge within the subgraph. The identified edge then serves as the criterion to decide which worker is assigned the task of processing that subgraph, leading to the relinquishment of the task by all other workers, without the need for inter-work coordination.

\subsubsection{Topological Attribute Separation Structure for Multi-Subgraph Optimization}
Since the system runs entirely in memory, and subgraph enumeration often requires huge space to maintain intermediate data, we develop a series of techniques to minimize memory consumption. Considering that multiple subgraphs will share the same node or edge, we propose a topological attribute separation structure to minimize communication and memory usage. Further, for a subgraph that has reached the maximum size, we output it immediately to relieve memory constraints, since it does not need to be expanded and propagated any longer. To conclude each iteration, we clear the useless subgraphs of each worker to prevent repeated propagation. Note that our method is able to support multi-edges between two nodes as well as multi-attributes on nodes and edges, which are not supported by most existing graph pattern mining methods.

\subsection{Two-staged Pattern Mining}

\subsubsection{Pattern Representation Mapping}
After obtaining potential subgraphs around each starting node, we need to perform an isomorphism test on a subset of all the structures to obtain the final candidate graph pattern. Nevertheless, exact matching based on graph isomorphism is computationally prohibitive for application to large-scale graph datasets.

\begin{figure}[t!]
  \centering
  \includegraphics[width=0.9\linewidth]{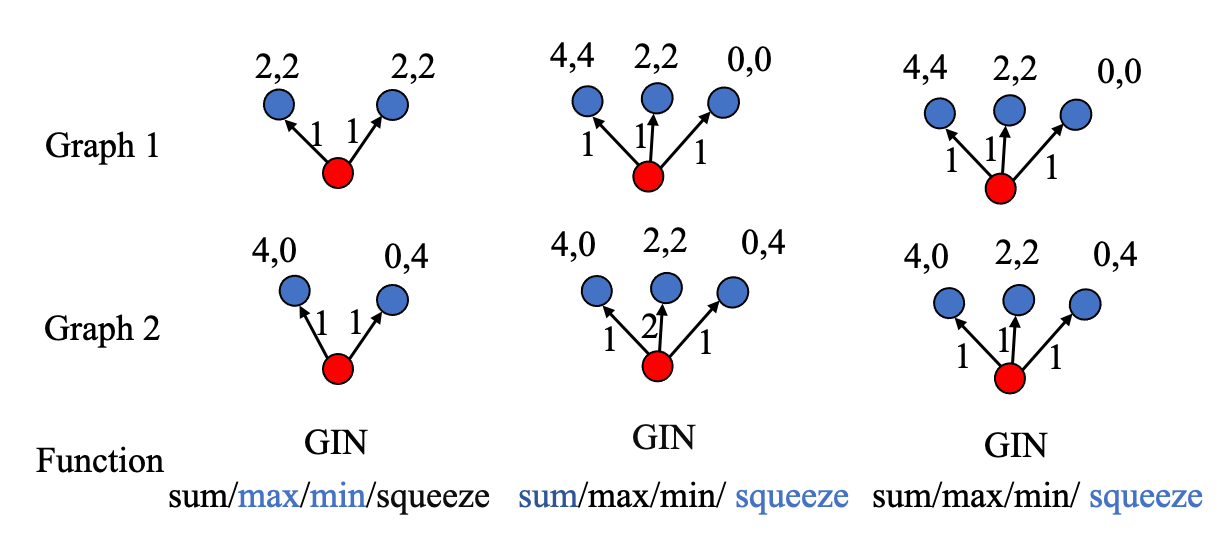}
  \vspace{-0.3cm}
  \caption{Examples of graph isomorphism test. GIN fails to handle the graph isomorphism test between the above graph instances. Functions labeled in blue can differentiate the graph 1 and the graph 2. Using the left figure as an example, when performing neighborhood aggregation for the red node, the \texttt{sum} and \texttt{squeeze} aggregators both generate an identical neighborhood embedding vector (4, 4) from the two blue nodes, which does not allow for differentiation between graphs 1 and 2. In contrast, the \texttt{max} aggregator yields a neighborhood embedding vector of (2, 2) for graph 1 and (4, 4) for graph 2. Similarly, the \texttt{min} aggregator produces embedding vectors of (2, 2) for graph 1 and (0, 0) for graph 2. Clearly, both the \texttt{max} and \texttt{min} aggregators are capable of discerning the differences between the two graphs. For the scenarios depicted in the middle and right figures, a comparable analytical approach can be applied.}
  \label{fig.egin_advantage}
  \vspace{-0.3cm}
\end{figure}

To circumvent this issue, we employ the Graph Isomorphism Network (GIN)~\cite{gin2019Xu}, whose efficacy has been equated with the Weisfeiler-Lehman (WL) isomorphism test~\cite{weisfeiler1968reduction}, as an approximate matching technique to diminish computational demands. It should be noted, however, that GIN encounters limitations when addressing the graph isomorphism problem~\cite{Principal2020Gabriele}, specifically in scenarios involving nodes with high-dimensional attributes or edges that bear attributes. As shown in Fig.~\ref{fig.egin_advantage}, unique graph schema may be erroneously mapped to an identical representation space when utilizing GIN, thereby rendering the isomorphism test for graphs unfeasible. Consequently, we introduce a novel architecture referred to as the Edge-Involved Graph Isomorphism Network (EGIN), designed to project the representations of the enumerated subgraphs into a high-dimensional representation space, specifically tailored for graph isomorphism tasks with high-dimensional attributes. Firstly, we integrate the edge information, including features and directions, into the GIN's message-passing mechanism.

\begin{equation}
\small
	\label{eq:node_encoder}
        \mathbf{h}_{i}^{(k)} = MLP^{(k)}_{n}((1+\epsilon^{(k)})\mathbf{h}_{i}^{(k-1)}+\sum_{v_j \in \mathcal{N}(v_i)}MLP_0(\mathbf{h}_{j}^{(k-1)}||\mathbf{f}_{e_{ij}}^{k-1}))
\end{equation}
where $\mathbf{h}_{i}^{(k)}$ is the intermediate representations of node $v_i$ in the $k$-th layer and $v_j \in \mathcal{N}(v_i)$ denotes the set of first-order neighboring nodes of the node $v_i$. $MLP(\cdot)$ denotes the multi-layer perceptron network, '$||$' is the concatenation operator and $\epsilon^{(k)}$ is a constant. The edge embedding of $\mathbf{f}_{e_{ij}}^{k}$ is calculated as
\begin{equation}
	\label{eq:edge_encoder}
        \mathbf{f}_{e_{ij}}^{k} = MLP_e^{(k)}(\mathbf{f}_{e_{ij}}^{k-1}||\mathbf{h}_{i}^{(k-1)}+\mathbf{h}_{j}^{(k-1)})
\end{equation}

Secondly, we can obtain the subgraph embedding $z_g$ of the subgraph $g$ with the $K$-th aggregation layer as follows.
\begin{equation}
    z_g = READOUT( h_i^{(K)} |  v_i \in \mathcal{V}_g )
\end{equation}
Simply using \texttt{sum} as \texttt{READOUT} is not able to distinguish the differences of high-dimensional attributed nodes, as shown in Fig.~\ref{fig.egin_advantage}. In this regard, we also introduce the aggregation operations of \texttt{max}, \texttt{min}, and \texttt{squeeze}, where \texttt{squeeze} refers to squeezing the features of multi-dimensional nodes into one-dimensional feature vector by employing a summation function. Finally, the subgraph embeddings produced by each of the four operators are concatenated to construct a composite representation for the current subgraph pattern.


The final merging of subgraph representations yields the representations of all candidate graph patterns that exist, and the set of nodes contained in each graph pattern can be computed by comparing whether the graph pattern's representation is the same as the representation of a subgraph under the node. The parameters of the EGIN network do not necessarily need training and can be initialized by ensuring the linear transformations in MLPs meet the injection property. This way, we can get identical representations for the same graph patterns and roughly get different representations for different graph patterns.

Through the method delineated above, we can obtain a collection of subgraph representations corresponding to all potential graph patterns, wherein representations of identical subgraphs should be precisely equivalent. Consequently, by comparing the representation of a graph pattern with the representations of subgraphs beneath individual nodes, the node set that each graph pattern encompasses can be computed. In our proposed methodology, it is essential for the parameters of the EGIN to exhibit a one-to-one (injective) mapping property.
This property guarantees that identical graph patterns are mapped to congruent representations, whereas disparate graph patterns are mapped to distinguishable representations.

Grounded in the principles of the universal approximation theorem~\cite{Hornik1989Multilayer,Hornik1991Approximation}, the utilization of an MLP within the framework of the EGIN is employed to ensure the injective mapping of graph representations. The EGIN model constitutes an augmentation of the GIN model, which has been previously established as possessing an equivalent level of discriminative capacity as the Weissfeiler-Lehman (WL) test for graph isomorphism with respect to homogeneous graphs. Given that EGIN integrates edge features into the original GIN structure while preserving the MLP's injective nature, it is posited that EGIN retains the discriminative prowess of its predecessor, thereby enabling the identification of distinctive graph patterns. Refer to~\cite{gin2019Xu} for a similar analysis.

\subsubsection{Two-staged Mining}
In terms of pattern representation mapping, the high-dimensional feature spaces associated with nodes and edges result in the magnitude of the final graph pattern being exponentially contingent upon the dimensionality of these features. In practice, graph patterns typically exhibit a long-tailed distribution, where numerous low-support patterns contribute to an inflated computational workload, leading to significant memory redundancy and inefficiency.

To ameliorate these computational concerns, we propose a two-stage pattern mining scheme, where only the features of nodes are used in the first stage for the pattern representation mapping task, and then the low-support graph patterns are pruned (according to the principle of anti-monotonicity, the support of graph patterns obtained by expanding the attributes on the edges of the low-support graph patterns will be even lower). Subsequently, in the mining's second stage, we perform the graph representation mapping task again based on the subgraphs hit by the top $p\%$ of supported patterns (by default $p$ is set to 10 to realize an empirically good trade-off between efficacy and efficiency), where the features on the edges are introduced for pattern merging, resulting in the final set of risky candidate patterns. Implementing this tiered framework dramatically streamlines the computational process, reducing the original complexity from $O(N\times E)$ to a more manageable $O(N+E)$.

In practical implementations, the prevalence of anomalous nodes is typically much lower than that of normal nodes within the majority of graphs. Therefore, it is operationally efficient to first mine for risk patterns $P$ on anomalous nodes, and then to calculate the support of these patterns on normal nodes. This strategy not only prevents redundant computations over a large set of normal nodes but also provides a measure of support for risk patterns across both anomalous and normal samples, which is essential for the subsequent assessment of risk patterns.

\subsection{Pattern Risk Assessment}
In the context of financial security, it is imperative for risk management systems to effectively differentiate between normal and anomalous entities within the trading network. To accomplish this, the system should identify risk graph patterns that are prevalent among anomalous nodes (indicative of potential risk) but are notably absent or rare among normal nodes. This requirement deviates from the aims of traditional frequent graph pattern mining, which generally seeks to find patterns that occur commonly across the entire graph without particular focus on anomaly detection. To quantify the efficacy of such discriminative patterns, we introduce a novel evaluation metric termed the Pattern Risk Score ($R_s$), specifically tailored for financial risk analysis tasks. This metric aids in the assessment of a pattern's reliability and relevance in identifying financial risks.

For a given pattern $P_i$, suppose there exists a set of nodes with associated binary labels $y_{v_i}\in \mathcal{Y}$ in the historical data, where each label signifies the node's status as either normal ($y_{v_i}=0$) or abnormal ($y_{v_i}=1$). We first calculate the support counts $s_{y=1}^{P_i}$ and $s_{y=0}^{P_i}$, representing the support of $P_i$ among abnormal and normal nodes, respectively.

Subsequently, we compute the precision of the pattern ($pre(P_i)$) as the ratio of its support among abnormal nodes to its total support across both normal and abnormal nodes, formalized as:
\begin{equation}
 pre(P_i) = \frac{s_{y=1}^{P_i}}{s_{y=1}^{P_i} + s_{y=0}^{P_i}} 
\end{equation}
The recall of the pattern ($re(P_i)$) is measured as the proportion of abnormal nodes that support the pattern relative to the total number of abnormal nodes, expressed as:
\begin{equation}
 re(P_i) = \frac{s_{y=1}^{P_i}}{\sum_{y_i=1} y_i} 
\end{equation}

To synthesize precision and recall, reflecting a pattern’s overall effectiveness, we calculate the Pattern Risk Score $R_s(P_i)$ analogously to the F1 score:
\begin{equation}
\label{eq:risk_score}
R_s(P_i) = \frac{ 2 \times pre(P_i) \times re(P_i)}{pre(P_i) + re(P_i)},
\end{equation}
where $R_s(P_i)$ balances the trade-off between precision and recall. A higher $R_s$ signifies a more reliable pattern in distinguishing financial risks. The pattern risk score quantifies the reliability of graph patterns as indicators of financial risk. By prioritizing graph patterns with high $R_s$ scores, financial institutions can focus on scrutinizing transactions or nodes that are most likely associated with fraudulent activities, ensuring proactive risk mitigation and regulatory compliance. It essentially translates the abstract concept of network anomalies into actionable intelligence that can safeguard financial operations.

\begin{table}[t]
 \small
 \caption{Statistics of datasets.}
	\label{tab:datasets}
\resizebox{0.95\linewidth}{!}{
	\begin{tabular}{lccc}
		\toprule
		 \textbf{ }     & Industrial dataset1(\textbf{M1})  	&Industrial dataset2(\textbf{M2})	 & Industrial dataset3(\textbf{M3})	  \\ 
		\midrule
            \#Nodes                  & 7,743              & 26,776            & 54,411,161                            \\
		  \#Edges                & 10,743            & 468,263           & 130,862,451                         \\
            \#{Node features}           & {5}               & {10}              & {21}                             \\
		  \#{Edge features}           & {2}               & {6}              & {5}                             \\
            \#{Anomalies}            & {203}                & {1,101}               & {28,165}                              \\
            {Pos. label meaning}    &{fraudster}     &{fraudster}         &{fraudster}          \\
		\bottomrule
	\end{tabular}
 }      
\end{table}

\section{Experiments}

In this paper, we validate the effectiveness of our proposed methodology by conducting experiments on three industrial datasets of varying magnitudes. Specifically, the datasets consist of a financial transaction network derived from the industrial platform, wherein the associated labels demarcate the users as either legitimate or fraudulent. For all evaluated tasks and datasets, the graph data were organized in accordance with the chronological sequence of transactions. Subsequently, a uniform temporal division was applied across all datasets: the initial $70\%$ of the time-ordered data was allocated for the mining of risk patterns, and the concluding $30\%$ was reserved for testing purposes. The statistics for the three datasets are summarized in Table~\ref{tab:datasets}. It should be noted that our choice of risk patterns is based on a thorough analysis of historical data, and we've used data from different time periods to prevent overfitting.
Herein, we would like to emphasize that we can not provide additional details regarding the datasets and the specific features of nodes and edges, and are unable to open-source our code, due to confidentiality constraints. Disclosing these information could potentially compromise the integrity of our risk control system.

\begin{figure}[t!]
  \centering
  \includegraphics[width=\linewidth]{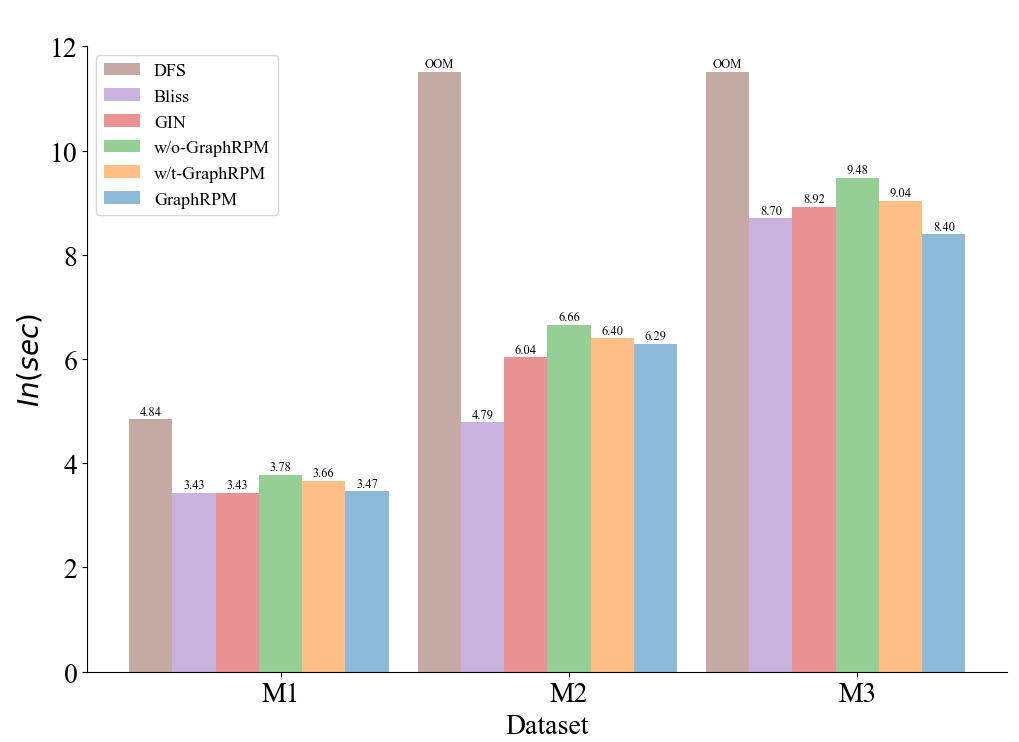}
  \caption{Overheads of different methods in terms of running time for different scale datasets.}
  \label{fig.efficiency}
  \vspace{-0.3cm}
\end{figure}

First, we validate the operational efficiency of our proposed method compared to the Exact-matching method, the Bliss~\cite{bliss2007} method, the GIN~\cite{gin2019Xu} method, w/o-GraphRPM (GraphRPM without enumeration optimization and two-stage mining optimization), and w/t-GraphRPM (GraphRPM without two-stage mining). The Exact-matching method is achieved by recursively comparing the connection and attributes of each node and edge in two subgraphs. The Exact-matching method, the Bliss method, and the GIN method are based on the results of subgraph enumeration optimization to do subgraph matching to verify the efficiency of EGIN. The experiments were run on a 32-node cluster with two 8-core Intel Xeon 8269CY CPUs and 16GB of DDR4 RAM per node for a fair comparison. The runtime of each method on different scale datasets is shown in Fig.~\ref{fig.efficiency}. Through the experiments, we find that the Exact-matching method is the least efficient and encounters the out-of-memory (OOM) error on middle-scale and large-scale datasets. The running time of our method is in the same order of magnitude as Bliss and GIN, but both of the latter can only use one-dimensional feature labels on nodes, which makes them unable to be applied to industrial attributed graphs. Furthermore, comparing w/o-GraphRPM and w/t-GraphRPM demonstrates that by using the subgraph enumeration optimization scheme and the two-stage mining framework, the running time of GraphRPM is reduced by $3\times$ and $2\times$ on the large-scale dataset, respectively. Overall, GraphRPM enables the efficient implementation of pattern mining on industrial-attributed graphs.

\begin{table}[t!]
  \caption{Abnormal node identification task results, where the reported metric is the pattern risk score, precision (\%), and recall(\%).}
  \label{tab:overall}
  \resizebox{\linewidth}{!}
  {
    \begin{tabular}{cl|cc|cc|cc}
      \toprule
      \multirow{2}{*}{\textbf{$\tau$}} & \multirow{2}{*}{\textbf{Methods}} & \multicolumn{2}{c}{\textbf{M1}} & \multicolumn{2}{c}{\textbf{M2}}   & \multicolumn{2}{c}{\textbf{M3}}                                                                                                                            \\
      \cmidrule{3-8}
                                                  &                                   & $\mathbf{R_s}$              & \textbf{pre/recall}                   & $\mathbf{R_s}$              & \textbf{pre/recall}           & $\mathbf{R_s}$           & \textbf{pre/recall}            \\
      \midrule
      \multirow{5}*{1}
                                                  & EXPERT                  &   0.515 &  97.1\%  / 35.0\%     &   0.165 & 95.3\%   /   9.1\%       &    0.126 & 91.4\% /  6.7\%          \\
                                                  & Exact-matching                     &  \textbf{0.596} &  84.6\%   /  46.1\%            &       -     &     -        & -           & -           \\
                                                  & Bliss                      &  0.547 &  55.3\%   /       54.2\%        &   0.266 &   21.8\%  /    34.2\%        &  0.166 & 10.8\%   / 36.1\%          \\
                                                  & GIN                             &  0.523 & 51.6\%  / 53.1\%      & 0.325 & 37.5\% / 28.7\%  & 0.202 & 16.9\% / 25.2\% \\
                                                  & GraphRPM                       & \textbf{0.596} & 84.6\%   /  46.0\%   & \textbf{0.377} & 81.3\%  / 24.56\% & \textbf{ 0.307} & 77.4\% / 19.1\% \\

      \midrule
      \multirow{5}*{10}
                                                  & EXPERT                      &    0.773 & 95.7\%   /  65.0\%    & 0.405 &  90.3\%  / 26.1\%     &    -   &    -     \\
                                                  & Exact-matching                     &       \textbf{0.809} & 80.4\% /    81.4\%             &        -    &      -       & -           & -          \\
                                                  & Bliss                      &   0.536 &   39.3\%    /  84.5\%   &   0.297 &  19.2\%  / 66.1\%  &   0.145& 8.5\% / 49.2\%      \\
                                                  & GIN                             &  0.670 & 59.1\% / 77.4\%    &  0.369 & 26.4\% / 61.3\%  &  0.219 & 14.9\% /  42.1\%\\
                                                  & GraphRPM                             &  0.804 &  79.4\%   / 81.6\%  &  \textbf{0.528} & 74.1\% / 41.1\% &\textbf{0.465} & 68.3\% /  35.3\% \\

      \midrule
       \multirow{5}*{100}
                                                  & EXPERT                         & -              & -             & -          & -           & -           & -           \\
                                                  & Exact-matching                     &   \textbf{0.716} &  58.9\%   /   91.3\%         &          -  &     -      & -           & -          \\
                                                  & Bliss                      &    0.475 & 31.7\%   /  94.8\%    & 0.228 &   13.3\%  /  82.1\%   &   0.061 &    3.1\% / 72.1\%        \\
                                                  & GIN                    &  0.603 & 44.6\%  / 93.2\%    & 0.297 & 18.9\% / 68.9\%   & 0.178 &10.2\%  / 66.1\%  \\
                                                  & GraphRPM                             & 0.706 & 58.0\%  / 90.2\%    &  \textbf{0.621} & 65.9\% / 58.8\% & \textbf{0.560} & 59.7\% / 52.8\% \\

      \bottomrule
    \end{tabular}
  }
\end{table}

Table~\ref{tab:overall} shows the performance of the different methods for the identification task on the industrial graphs, where $\tau$ specifies the number of header risk patterns selected based on the pattern risk score. \texttt{EXPERT} refers to manually constructed risk patterns by experts, which cannot produce a large number of risk patterns due to high labor costs. Among them, expert patterns tend to have high accuracy rates, which are finely constructed from expert experiences.
However, due to the high efforts, it is difficult to expand and apply to various businesses. GraphRPM significantly outperforms the Bliss and GIN methods that use only one-dimensional features, especially on the largest dataset M3, where the risk score performance improves by 0.49 and 0.38, respectively, when using 100 patterns. Moreover, compared to the Exact-matching method, our method performs similarly on small-scale datasets but can operate on large-scale attributed graphs.

\section{Deployment}

\begin{figure}[t!]
  \centering
  \includegraphics[width=\linewidth]{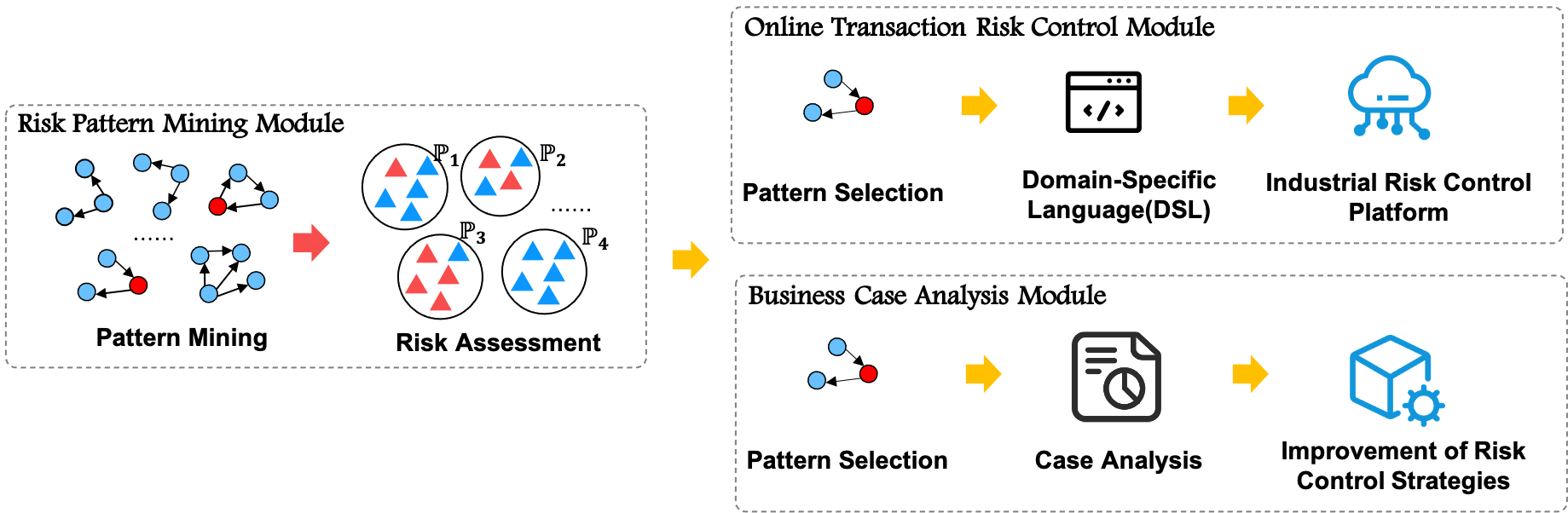}
  \caption{The deployment process of GraphRPM within the context of financial transaction scenarios.}
  \label{fig.deployment}
\end{figure}

In this section, we will elucidate the deployment process of GraphRPM within the context of financial transaction scenarios. As illustrated in Fig.~\ref{fig.deployment}, the deployment is segmented into three distinct modules: Risk Pattern Mining based on historical data, Online Transaction Risk Control based on identified risk patterns, and Business Case Analysis also based on identified risk patterns.

\textbf{Risk Pattern Mining Module}
This module operates by employing the GraphRPM method to mine patterns from historical transaction data. Subsequent to pattern risk assessment, the top risk patterns are selected for utilization in downstream tasks. These patterns are derived through an analysis of historical data, focusing on identifying key structures that have been associated with risky behavior in past transactions. Furthermore, considering the timeliness required for risk control, we update these patterns on a daily basis.

\begin{figure}[t!]
  \centering
  \includegraphics[width=0.7\linewidth]{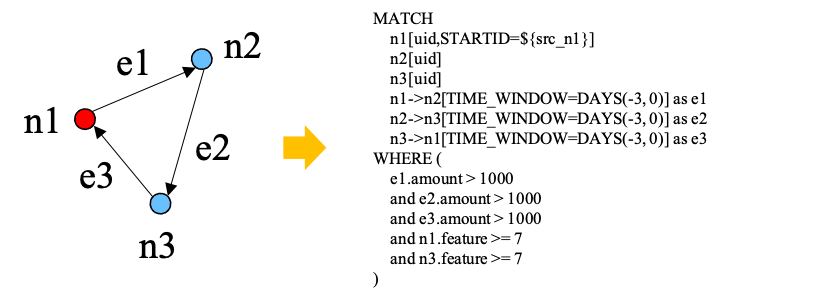}  \caption{An example of a graph pattern corresponding to domain-specific languages.}
  \label{fig.dsl}
\end{figure}

\textbf{Online Transaction Risk Control Module}
The selected risk patterns can be applied to online risk control. We encode these patterns into a domain-specific language (DSL) and deploy them to an industrial risk control platform. For example, Fig.~\ref{fig.dsl} demonstrates a snippet of DSL that describes a triangular structure. Every ongoing transaction is then scrutinized in real-time against these risk patterns. If a current transaction fully matches a risk pattern within its adjacent graph data, certain control measures can be enforced to restrict the completion of the transaction; otherwise, the transaction is permitted to proceed.

\textbf{Business Case Analysis Module}
Utilizing risk patterns for retrospective analysis, key graph structures can be extracted from past fraud cases. This assists business analysts in dissecting the tactics used by fraudulent operators, thereby enhancing the efficiency of the analytical process. By applying identified risk patterns to past cases, we can gain insights into the modus operandi of fraudsters and potentially anticipate future fraudulent schemes.

A critical discussion point is the adversarial nature of transactional risks. Once fraudulent actors can no longer exploit identified patterns due to enhanced risk control, they may devise new methods or channels to perpetrate fraud, circumventing current risk control patterns. Therefore, it is essential to perform periodic updates to the risk pattern. Regularly applying the Risk Pattern Mining Module to new data samples is necessary to unearth a new set of effective risk patterns for ongoing risk control measures.

\section{Conclusion}

Pattern mining on large-scale attributed graphs is always a major challenge in the field of data mining and machine learning. In this study, we introduce GraphRPM, an innovative framework that integrates a subgraph isomorphism algorithm powered by graph neural networks with an architecture optimized for computational efficiency. GraphRPM is designed to find and evaluate risk graph patterns automatically on large attributed graphs, helping to detect clusters of risky behaviors while reducing manual inspection costs, and has been \textit{deployed in production for more than one year in a diversity of business scenarios}. Through comprehensive experimentation on three diverse datasets of varying sizes, we establish that GraphRPM efficaciously addresses the challenges of pattern mining in large-scale attributed graphs prevalent in industrial contexts, underscoring its substantial value for industrial applications.

\begin{credits}
\subsubsection{\ackname}
We acknowledge that the datasets are only used for academic research, and do not represent any real business situation. They are desensitized and encrypted, do not contain any Personal Identifiable Information and were destroyed after the experiments. Adequate data protection was carried out during the experiment to prevent the risk of data copy leakage. Sheng Tian, Xintan Zeng and Yifei Hu are co-first authors with equal contributions. Baokun Wang and Yongchao Liu are joint corresponding authors.

\subsubsection{\discintname}
The authors have no competing interests to declare that are
relevant to the content of this article.
\end{credits}
%
%
%
%

\begin{thebibliography}{10}
\providecommand{\url}[1]{\texttt{#1}}
\providecommand{\urlprefix}{URL }
\providecommand{\doi}[1]{https://doi.org/#1}

\bibitem{Principal2020Gabriele}
Corso, G., Cavalleri, L., Beaini, D., Li{\`{o}}, P., Velickovic, P.: Principal
  neighbourhood aggregation for graph nets. In: Larochelle, H., Ranzato, M.,
  Hadsell, R., Balcan, M., Lin, H. (eds.) In: NeurIPS. (2020)

\bibitem{Migration2009xiaoxi}
Du, X., Jin, R., Ding, L., Lee, V.E., Jr., J.H.T.: Migration motif: a spatial -
  temporal pattern mining approach for financial markets. In: IV, J.F.E.,
  Fogelman{-}Souli{\'{e}}, F., Flach, P.A., Zaki, M.J. (eds.) In: SIGKDD. (2009).

\bibitem{GRAMI2014Mohammed}
Elseidy, M., Abdelhamid, E., Skiadopoulos, S., Kalnis, P.: {GRAMI:} frequent
  subgraph and pattern mining in a single large graph. Proc. {VLDB} Endow.
  \textbf{7}(7),  517--528 (2014)

\bibitem{gonzalez2012powergraph}
Gonzalez, J.E., Low, Y., Gu, H., Bickson, D., Guestrin, C.: Powergraph:
  Distributed graph-parallel computation on natural graphs. In: OSDI. pp.
  17--30 (2012)

\bibitem{Output2009Mohammad}
Hasan, M.A., Zaki, M.J.: Output space sampling for graph patterns. Proc. {VLDB}
  Endow.  \textbf{2}(1),  730--741 (2009)

\bibitem{Hornik1991Approximation}
Hornik, K.: Approximation capabilities of multilayer feedforward networks.
  Neural Networks  \textbf{4}(2),  251--257 (1991)

\bibitem{Hornik1989Multilayer}
Hornik, K., Stinchcombe, M.B., White, H.: Multilayer feedforward networks are
  universal approximators. Neural Networks  \textbf{2}(5),  359--366 (1989)

\bibitem{bliss2007}
Junttila, T.A., Kaski, P.: Engineering an efficient canonical labeling tool for
  large and sparse graphs. In: Proceedings of the Nine Workshop on Algorithm
  Engineering and Experiments. {SIAM} (2007)

\bibitem{lin2018shentu}
Lin, H., Zhu, X., Yu, B., Tang, X., Xue, W., Chen, W., Zhang, L., Hoefler, T.,
  Ma, X., Liu, X., Zheng, W., Xu, J.: Shentu: Processing multi-trillion edge
  graphs on millions of cores in seconds. In: SC18. pp.
  706--716 (2018)

\bibitem{liu2022eriskcom}
Liu, F., Li, Z., Wang, B., Wu, J., Yang, J., Huang, J., Zhang, Y., Wang, W.,
  Xue, S., Nepal, S., Sheng, Q.Z.: Eriskcom: An e-commerce risky community
  detection platform. The VLDB Journal  \textbf{31}(5),  1085–1101 (2022)

\bibitem{liu2022graphtheta}
Liu, Y., Li, H., Zhang, G., Zeng, X., Li, Y., Huang, B., Zhang, P., Li, Z.,
  Zhu, X., He, C., Chen, W.: Graphtheta: A distributed graph neural network
  learning system with flexible training strategy. arXiv preprint
  arXiv:2104.10569  (2022)

\bibitem{luo2023faf}
Luo, Y., Wang, G., Liu, Y., Yue, J., Cheng, W., Fei, B.: Faf: A risk detection
  framework on industry-scale graphs. In: CIKM. p.
  4717–4723 (2023)

\bibitem{Community2017Seyed}
Moosavi, S.A., Jalali, M., Misaghian, N., Shamshirband, S., Anisi, M.H.:
  Community detection in social networks using user frequent pattern mining.
  Knowl. Inf. Syst.  \textbf{51}(1),  159--186 (2017)

\bibitem{Subgraph2022Nguyen}
Nguyen, L.B.Q., Zelinka, I., Sn{\'{a}}sel, V., Nguyen, L.T.T., Vo, B.: Subgraph
  mining in a large graph: {A} review. WIREs Data Mining Knowl. Discov.
  \textbf{12}(4) (2022)

\bibitem{talukder2016a}
Talukder, N., Zaki, M.J.: A distributed approach for graph mining in massive
  networks. Data Min. Knowl. Discov.  \textbf{30}(5),  1024–1052 (2016)

\bibitem{teixeira2015arabesque}
Teixeira, C.H.C., Fonseca, A.J., Serafini, M., Siganos, G., Zaki, M.J.,
  Aboulnaga, A.: Arabesque: A system for distributed graph mining. In: SOSP. p.
  425–440 (2015)

\bibitem{weisfeiler1968reduction}
Weisfeiler, B., Leman, A.: The reduction of a graph to canonical form and the
  algebra which appears therein. nti, Series  \textbf{2}(9),  12--16 (1968)

\bibitem{gin2019Xu}
Xu, K., Hu, W., Leskovec, J., Jegelka, S.: How powerful are graph neural
  networks? In: ICLR. (2019)

\bibitem{yan2002gspan}
Yan, X., Han, J.: gspan: graph-based substructure pattern mining. In: ICDM. pp. 721--724
  (2002)

\bibitem{tfsm2023yuan}
Yuan, L., Yan, D., Qu, W., Adhikari, S., Khalil, J., Long, C., Wang, X.:
  {T-FSM:} {A} task-based system for massively parallel frequent subgraph
  pattern mining from a big graph. Proc. {ACM} Manag. Data  \textbf{1}(1),
  74:1--74:26 (2023)

\bibitem{Efficiently2005Mohammed}
Zaki, M.J.: Efficiently mining frequent trees in a forest: Algorithms and
  applications. {IEEE} Trans. Knowl. Data Eng.  \textbf{17}(8),  1021--1035
  (2005)

\bibitem{zhu2016gemini}
Zhu, X., Chen, W., Zheng, W., Ma, X.: Gemini: A computation-centric distributed
  graph processing system. In: OSDI. pp. 301--316 (2016)

\end{thebibliography}

\end{document}